\documentclass[10pt,twocolumn,letterpaper]{article}

\pdfoutput=1
\usepackage{iccp}
\usepackage{times}
\usepackage{epsfig}
\usepackage{graphicx}
\usepackage{amsmath}
\usepackage{amssymb}

\newcommand{\Section}[1]{\vspace{-6pt}\section{\hskip
-1em.~~#1}\vspace{-4pt}}
\newcommand{\SubSection}[1]{\vspace{-6pt}\subsection{\hskip
-1em.~~#1}\vspace{-4pt}}

\iccpfinalcopy 

\def\iccpPaperID{26} 

\ificcpfinal\fi
\begin{document}

\title{\  Progressive versus Random Projections for Compressive Capture of Images, Lightfields and Higher Dimensional Visual Signals }

\author{Rohit Pandharkar\\
MIT Media Lab\\
75 Amherst St, Cambridge, MA
\and
Ashok Veeraraghavan\\
MERL\\
201 Broadway, Cambridge MA
\and
Ramesh Raskar\\
MIT Media Lab\\
75 Amherst St, Cambridge, MA
}

\maketitle

\begin{abstract}
Computational photography involves sophisticated capture methods. A new trend is to capture projection of higher dimensional visual signals such as videos, multi-spectral data and lightfields on lower dimensional sensors. Carefully designed capture methods exploit the sparsity of the underlying signal in a transformed domain to reduce the number of measurements and use an appropriate reconstruction method. Traditional progressive methods may capture successively more detail using a sequence of simple projection basis, such as DCT or wavelets and employ straightforward backprojection for reconstruction. Randomized projection methods do not use any specific sequence and use $L_{0}$ minimization for reconstruction.
In this paper, we analyze the statistical properties of natural images, videos, multi-spectral data and light-fields and compare the effectiveness of progressive and random projections. We define effectiveness by plotting reconstruction SNR against compression factor. The key idea is a procedure to measure best-case effectiveness that is fast, independent of specific hardware and independent of the reconstruction procedure. We believe this is the first empirical study to compare different lossy capture strategies without the complication of hardware or reconstruction ambiguity. The scope is limited to linear non-adaptive sensing. The results show that random projections produce significant advantages over other projections only for higher dimensional signals, and suggest more research to nascent adaptive and non-linear projection methods.

\end{abstract}

\Section{Introduction}
Computational photography involves sophisticated capture methods to capture high dimensional visual signals using invertible multiplexing of signals. This is achieved by careful capture time projection followed by sophisticated reconstruction. To reduce the number of measurements, a common strategy is to exploit the sparsity in a transformed domain. Let us consider the two projective signal capture approaches for exploiting the sparsity: progressive versus randomized sampling.

{\bf Progressive projections} Consider the single pixel camera \cite{ Singlepixel}. The successive basis from domains like DCT or Wavelet can be used as modulation patterns to progressively capture higher frequencies. Since signal energy is often compactly represented with the first few coefficients, recovering them is useful for reconstruction. $M$ measurements using these progressive projection patterns lead to recovery of first $M$ frequency coefficients within the basis chosen. The reconstruction is straightforward via weighted combination of each of the $M$ basis.This has been well-known technique for visual signal compression for a long time (JPEG \cite{275888} , MPEG \cite{1291239}, Transform coding \cite{537169}, Wavelet based compression \cite{WaveletTC}).

{\bf Random projections} Random projection based methods are often studied in the compressive sensing \cite{candes2006rup}\cite{DBLP:journals/tit/Donoho06} and have been applied to various acquisition problems in vision and graphics. Noteworthy among these are: Single pixel camera \cite{Singlepixel}, Compressed video sensing\cite{Drori_compressedvideo}, CS Light field capture \cite{Babacan:2009:667}, and Multispectral capture \cite{Gehm:07}.  This has spun a new array of techniques that observe linearly mixed random measurements (projections) and reconstructs the signal using compressive sensing based reconstruction algorithms ($L_{1}$ minimization). For signals that can be shown to be sparse in some basis, it has been shown that observing $M = cKlog(\frac{N}{K}) $ linear observations is sufficient to ensure exact recovery of such signals \cite{2006ssr}.

We aim to empirically answer the question: Are random projections based signal capture methods more effective than progressive projections for visual signals? This question, though important has not been answered concretely in the literature due to multiple challenges: (a) The first challenge is to define effectiveness. (b) The second challenge is to create a metric that is independent of a specific capture strategy or the sophistication of the (future) reconstruction algorithm. (c) The third challenge is to accommodate nascent adaptive projection methods in comparison. These factors make the discussion on sparsity exploiting methods very tortuous.
We address this using the following approach: (a) We compare effectiveness by computing the reconstruction SNR (signal to noise ratio) for a given compression factor. The approach is empirical and we employ large datasets for analysis. Because the datasets are large for analyzing the statistical properties, we also need a fast and efficient method for comparing effectiveness. (b) We use a metric based on Parseval's energy theorem to relate energy of signal to the energy of the coefficients. Then we use best-case of most recent empirically found recovery bounds to be partial to randomized projection approaches. (c) We limit ourselves to linear and non-adaptive projection methods which are mature and well understood. When the adaptive methods become more established, we agree that a new comparative analysis will need to be performed.
In this paper, we design a mechanism to compute the effectiveness for a variety of visual signals such as videos, multi-spectral data and lightfields.
\vspace{-0.1in}
\subsection{Motivation}
While sparsity has been exploited for image acquisition \cite{Singlepixel}, dual photography \cite{Sen:CS_DualPhoto:2009}, reflectance field capture \cite{DBLP:journals/tog/PeersMLGMRD09} and face recognition \cite{wright2009robust}, there has been little attention paid to the comparison of progressive projections and random projections. The primary analysis in this paper indicates that for images and videos, progressive projection based capture and reconstruction techniques perform comparable to that of random projection based capture and reconstruction. Nevertheless, the analysis also shows that for higher dimensional computational photographic signals such as multispectral data and light-fields, sparsity plays a larger role and techniques that can recover highest magnitude frequency coefficients may be better equipped to tackle these problems, instead of techniques that progressively recover first few coefficients.

\begin{figure}
\centering
\includegraphics[width=8.5cm]{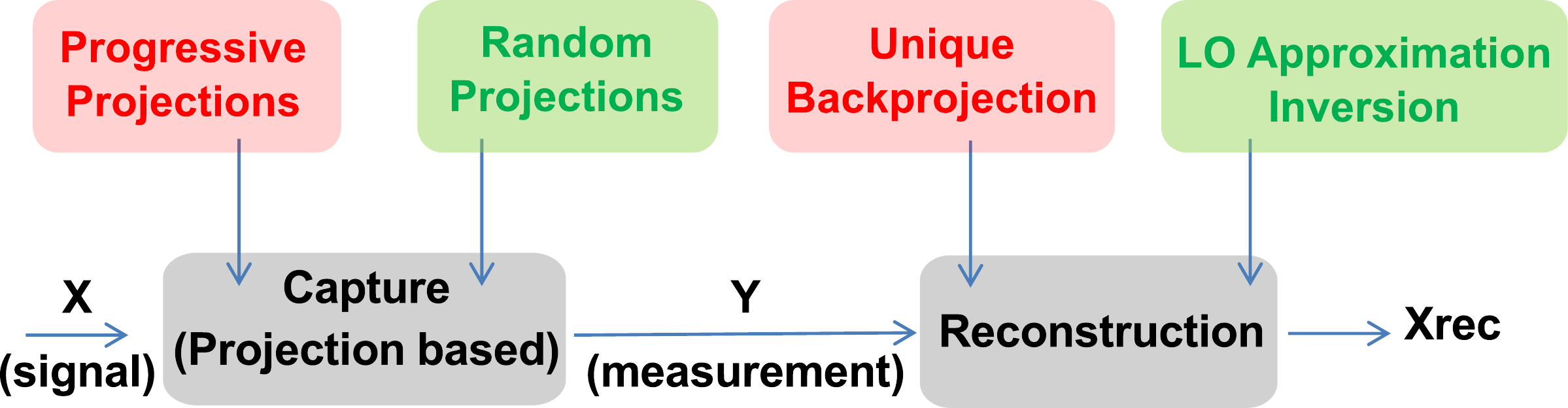}
\caption{ Pipeline for Signal capture and reconstruction using Progressive Transform Coding (TC, Red) or Randomized sparsity coding (SC4,Green).
}
\label{fig:newplot1}
\vspace*{-3ex}
\end{figure}

{\bf Objective and scope:} The objective of this paper is to empirically evaluate the relative merits of these sampling techniques. We statistically analyze the compressibility and sparsity structure of visual signals. In an attempt to be invariant to the choice of basis for progressive projection based capture techniques and the choice of basis for sparse signal approximation in random projection based capture, we try several basis functions including DCT, wavelet, Fourier and PCA basis for datasets and compare the best case results. Further, different reconstruction algorithms for compressive sensing basis pursuit, $L_1$ minimization, matching pursuit methods, etc each have a different performance for different kinds of data. It has been empirically noted that for all of these methods a minimum of $M=4K$ measurements are required to reconstruct the $K$-largest magnitude approximation of the sparse signal. This is generally considered an optimistic expectation. In an effort to remain independent of the numerical limitations of the reconstruction algorithms, we do not use any reconstruction algorithm, instead we study the energy of the approximation in $K=M/4$ highest magnitude coefficients, where M is the number of linear measurements. To be favorable to random projections, We:(a) ignore hardware limitations, (b) do not consider effect of quantization. However, the analysis in this paper is limited in the following ways:,we: (a) consider only linear measurements, (d) do not consider structural sparse representations \cite{DBLP:journals/corr/abs-0808-3572}, (e) do not consider adaptive methods like- learned dictionaries \cite{1553463} \cite{1772035} \cite{Weiss_learningcompressed} or overcomplete dictionaries \cite{1231494} or hybrid sensing techniques \cite{Ashok:10}
\vspace{-0.1in}
\subsection{Contributions}
\begin{itemize}
\item We devise a procedure to verify effectiveness of progressive as well as random projection based capture methods by comparing the reconstruction SNR for each compression factor. Importantly, the comparison method is fast and independent of capture hardware or reconstruction algorithm.
\item We analyze the sparsity, compressibility and energy compaction of several visual signals such as images, videos, multi-spectral imaging and light-fields.
\item We empirically demonstrate the analysis using a large dataset for multiple classes of visual signals.
\end{itemize}

\Section{Methodology}
Traditional signal processing and sampling theory begins with a continuous domain signal $x(t)$ that is then discretized at some sampling rate which is typically greater than the Nyquist rate. In this paper, we consider signals as discrete domain signals represented after they have been sampled and quantized. Further, we only consider a finite time horizon leading to a finite dimensional discrete vector. Let $x$ denote a finite dimensional discrete vector representing the original signal that needs to be compressed.
\begin{figure}
\centering
\includegraphics[width=8.5cm]{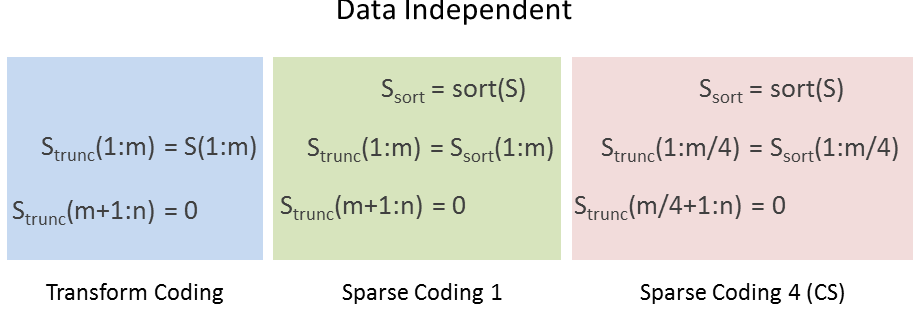}
\caption{ Pseudo codes for computations. The table shows compression techniques as implemented logically. 'S' is the representation of signal in frequency domain. Above pseudo codes signify the possible recoveries of frequency coefficients ($S_{trunc}$) through: Progressive projection capture (TC: Use of first M coefficients), Random projection capture and $L_{1}$ inversion (SC4: Use of highest magnitude M/4 coefficients) and Oracle capture (best possible: SC1: Use of M highest magnitude coefficients) all for $C=\frac{m}{n}$}
\label{fig:codenew}
\vspace*{-3ex}
\end{figure}

{\bf Projective capture based comparison:} Our goal in this paper is to devise a scheme for comparison of signal capture using progressive transform projections that have been devised with the ideas of energy compaction in mind. Traditional energy compaction techniques, when used directly in a hardware device observe linear combinations which are directly the dot products with the basis functions. When compression is required they observe far fewer dot products, i.e., dot products only with the first few frequency basis elements (it is well documented that signals usually have more energy in the first few coefficients). The newer emerging techniques of sparse representation and compressive sensing, instead rely on the fact that the signal is sparse in some appropriate domain. In hardware this amounts to observing linear combinations (or dot products) of the signal with some random binary (or Gaussian) vectors. The best estimate of the signal is then obtained via an L1 optimization algorithm. In order to compare the effectiveness of these two techniques for the process of signal acquisition, we devise a scheme that accurately approximates the reconstruction quality of a large class of visual signals. Further, the technique we propose is device and algorithm independent in the sense that limitations of the hardware sensing device or the computational algorithms do not affect our results.

\SubSection{Choice of Basis}
We would like our analyses to be as basis independent as possible. In order to achieve this, we analyze the performance of each of the methods using a variety of commonly used data independent basis such as DCT, wavelet, Fourier etc. Further, in order to evaluate performance characteristics due to the use of data dependent basis, we also use bases obtained via principal component analysis on a separate dataset for each of the datasets on which we evaluate our results. Recently, much progress has been made in the field of dictionary learning, which is another data-dependent basis better suited to sparse representations. While, performance improvements over PCA maybe obtained via the careful use of dictionary learning methods, these methods are yet to reach a stage of maturity that allow easy empirical evaluations. Therefore, in this paper we restrict our attention to data-independent bases and to PCA basis.
\SubSection{Comparison of projections}
For the rest of this section, let us assume that we have a hardware device available with us, capable of observing $M$ measurements, where each measurement is a linear combination of the signal with some known vector. The hardware device is not restricted in the sense that it can
obtain a linear combination with any real valued vector. Similarly, we will assume that at the back-end we have computational capacity to reach the optimal solution that is feasible and are not restricted by computational constraints. (Pseudo codes in Figure\ref{fig:codenew}).

\textbf{Progressive Transform Coding (TC):} This method attempts to approximate the reconstruction performance that would be obtained while adapting traditional compression techniques to capture devices. In hardware, this would amount to obtaining direct linear combinations with the basis elements (in the order of increasing frequency). Reconstruction, would amount to a backprojection into the signal domain which again is a linear operation. We use the well known Parseval's theorem to approximate the reconstruction fidelity. When $M$ measurements are obtained, this corresponds to the energy captured by the first $M$  frequency basis elements, irrespective of the magnitude of these coefficients.
Signals can be represented using their projections onto the basis vectors as $s=[s_1 s_2 s_3 ......]^T = \Phi x$, where $s$ represents the transform domain representation of the signal.
The signal itself can be reconstructed from the basis coefficients as $\hat x = \Phi^{-1} s = B s $ since $\Phi$ is a full rank matrix.
This can also be written as a linear combination of the basis functions as
$x= \sum_1^{N} s_i B_i$,
where $B_i$ is the $i^{th}$ basis function.
Most common bases in which signals are represented include the discrete cosine transform (DCT), wavelet transform and the Fourier transform. Most of the times, first few components of the representation $s_1,s_2,s_3,....s_K$ contain most of the signal energy.
Thus a K-term approximation for the signal can be obtained via $x_{K-term} = \sum_1^{K} s_i B_i$.
This approximation usually captures significant amount of the signal energy. Thus, if each progressive projection pattern were made up of $s_i$ basis vectors, progressively observing 'M' measurements of the progressive projections would be an 'M' term approximation of the signal in respective basis.

\textbf{Oracle Sparsity Coding 1 (SC1):} The goal of this method is to obtain an upper bound for the performance of sparse coding and compressive sensing techniques. As in traditional random projection methods the capture hardware would amount to observing random linear combinations with either random binary or random Gaussian entries.
Traditional random projections based reconstruction predicts that if $M$ measurements are observed then one would be able to reconstruct $K$ highest magnitude basis weights where K is given by $M=K log(N/K)$. Thus, having observed $M$ measurements allows us to only reconstruct $K<M$ basis weights. In the best case scenario, where an oracle allows us knowledge of the support (i.e. which coefficients have highest magnitude-unavailable in any practical setup), one can reconstruct the M highest magnitude coefficients from the M measurements. Thus the energy captured by $M$ highest magnitude coefficients represents an upper bound on the reconstruction performance using random projections. Most practical methods would perform much poorer than this upper bound.

\textbf{Randomized SparsityCoding (SC4):} As described before,Randomized sparsity coding observes random linear combinations and we expect to reconstruct K highest magnitude coefficients when we solve an L1 optmization for reconstruction. Here we define that for $s= \phi x$, we measure Y as $y= \psi \phi x$, where S is sparse to some extent and $\psi$ is a random matrix that satisfies the restricted isometry property (RIP). It should be noted that the extent of sparsity in S is not exactly known here (as is the case in most of the visual signals). Here $x_{rec}$ is obtained by solving the linear system $y= \psi \phi x$ using $L_{1}$ minimization (basis pursuit).In theory $M= cKlog (\frac{N}{K})$ linear measurements are required where c is an unknown constant. But in practice an efficient and robust algorithm for reconstruction might be able to reconstruct $K=M/4$ highest magnitude coefficients. While , this estimate is also optimistic, it is realistic to expect that current and future algorithmic advances will allow us to reach this. This would amount to the energy captured by the highest magnitude $M/4$ coefficients and serves as a realistic best case performance of Randomized sparsity coding.
Thus, this process at its best possible inversion performance is equivalent to Oracle sparsity coding (SC1) but varies in the sense that for a use of M measurements only M/4 coefficients effectively contribute.
While all other methods explained hitherto are data independent (except the sparsity prior in Randomized sparsity coding), we also experiment with Principal Component Analysis-a data dependent compression technique .
\SubSection{Evaluation Metrics:}
To formally analyze the extent of compression achieved and the quality of reconstructed signals, following statistical parameters can be used.The elements in $s$ are the coefficients for describing the signal in transform domain.

{\bf Compression Factor (C):} We define the compression factor as the number of coefficients used out of the total number of maximum coefficients.
{\centering $C= \frac{M}{N}$}

{\bf Reconstruction SNR (SNR$_{rec}$):}
Reconstruction SNR is the Signal to Noise Ratio defined for the reconstruction quality. We define this as the ratio of original signal energy to the error of the reconstructed signal compared to original signal. If original signal is x and reconstructed signal is $x_{rec}$
\begin{center}SNR = 20 $log_{10}[\frac{||x||}{||(x-x_{rec})||}$] dB \end{center} From Parseval's energy theorem,
\begin{center}SNR = 20 $log_{10}[\frac{||s||}{||s-s_{trunc}||}$] dB \end{center}
This key idea allows us to compare SNRs directly in sparse coefficient domain without worrying about reconstructing $X_{rec}$ from $S_{trunc}$. Thus the method is fast and independent of capture hardware or reconstruction algorithm.

\begin{figure}
\centering
\includegraphics[width=8.5cm]{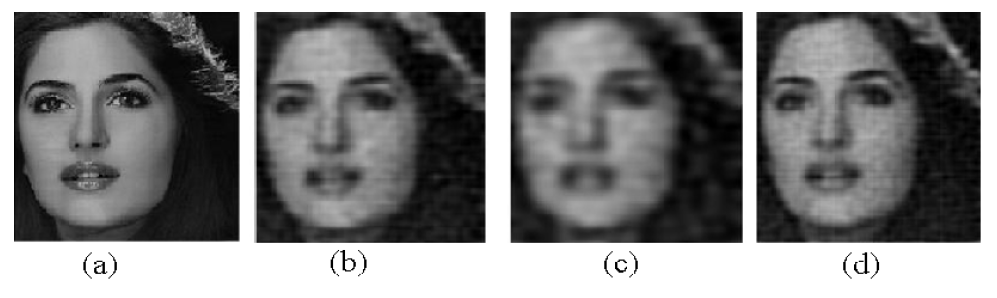}
\caption{ Example of compression of an image using competitive compression techniques with a constant compression factor of 0.02. (a) Original image (b) Compression using Progressive projection capture -TC (c) Compression using Random projections capture-SC4 (d) Compression using Oracle Capture-SC1. Performance order is: $SC1>TC>SC4$. This shows a case where Progressive projection performs better than Random projection.}
\label{fig:example}
\vspace*{-3ex}
\end{figure}

\Section{Datasets}
For carrying out an empirical analysis of how random projection capture (compressed sensing) performs in comparison with other compression techniques we do the experiments over competitive methods of signal compression and reconstruction.
We elaborate the competitive techniques used for signal compression in the signal processing literature and explain every technique for a generalized case of signal reconstruction (independent of dimensions: images, videos, multispectral data, light fields or signal parameters being compressed: light field spatial resolution, light fields angular resolution).
For generality, let us assume that we have total N measurements and for compression we use M measurements (C= m/n).
Also, let us assume that $s = [s_1 s_2 ... s_n]$ and $s_{trunc}= [s_{t1} s_{t2} .... s_{tn}]$.
We performed the SNR vs Compression factor analysis of following visual signals over a range of compression factors [0 to 1] for the following datasets:
(1) Images (2D) (a) Real images included green channel of 1000 random 2MP Flickr images resized to 256 $\times$ 256 pixels. (b) Cartoon images included green channel of 1000 random 2MP high quality cartoon images resized to 256$\times$256 pixels. (c) Face images included 4596 images from Yale B database \cite{GeBeKr01}. For PCA, the dataset was split into 
4096 test images for learning and 500 gallery images for operation.(2) Videos included 30 uncompressed videos resized to $64\times 64 \times 64$ pixels.(3) Multi Spectral Data included Columbia CAVE labs database, 25 scenes with 512 pixels$\times$ 512 pixels $\times$ 31 wavelengths. For PCA, cutout versions included 15 by 15 by 31 wavelength (= 6975) data points. (4)Light fields from (New)Stanford light field archive included
14 sets of 17 $\times$17 grid (289 views) with 1024 by 1024 resolution (resized to 256 $\times$ 256 pixels).
For more details see supplementary material

\begin{figure}
\centering
\includegraphics[width=7.5cm]{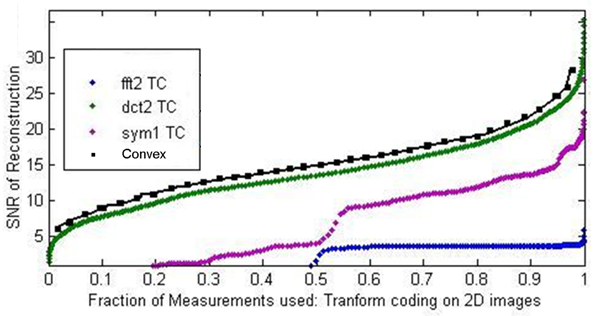}
\caption{ Individual performance plots. We compare the individual performance of wavelets, FFT and DCT for each compression method (TC, SC1, SC4) and plot the best performance curves for each. The black curves show best performance curves.}
\label{fig:Individual1}
\vspace*{-3ex}
\end{figure}

\begin{figure}
\centering
\includegraphics[width=7.5cm]{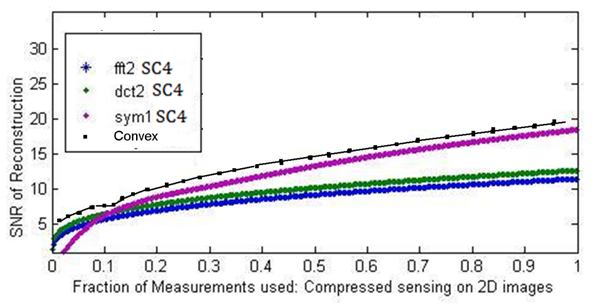}
\caption{ The plot shows how individual best performance curves are used to plot the final result}
\label{fig:Individual2}
\vspace*{-3ex}
\end{figure}

\Section{Results}
In order to perform the required comparison we analyzed the relative sparsity and energy compaction of visual signals in various data independent basis such as DCT, wavelet, Fourier etc and also the data-dependent PCA basis. We present the reconstruction performance Vs Compression factor plots for: Progressive projection capture (TC),Oracle capture (SC1) and Random projection capture (SC4).
For each of the three methods, i.e., TC,SC1 and SC4 we plot the SNR vs compression factor for several different choice of basis. In most cases, the bases that were chosen were DCT, Fourier, Haar Wavelet and Farras Wavelet \cite{DBLP:journals/tsp/AbdelnourS05}.
Once the separate performance plots for each of the individual bases were obtained, (as shown in Figure \ref{fig:Individual1}), then for each compression factor, we only retain the best performing basis.
Thus in essence, the performance plots we show are the convex hull (best case) of the performance plots for each of the individual bases as in figure \ref{fig:Individual2}.

\begin{figure}
\centering
\includegraphics[width=7.5cm]{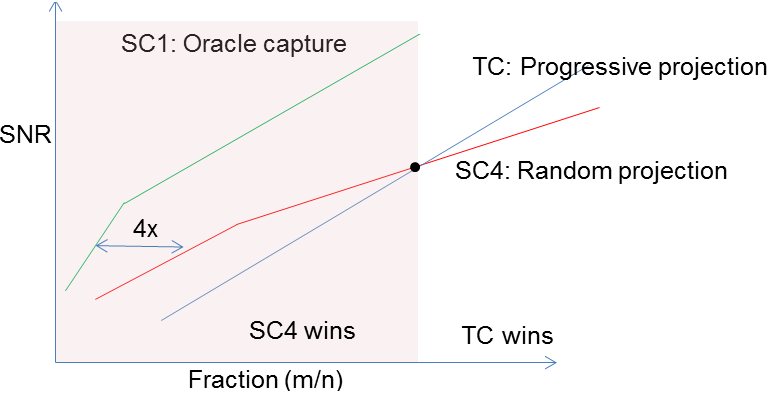}
\caption{How to interpret plots?: The approach in this paper is to validate Randomized sparsity coding (SC4) method against Progressive transform coding (TC). Horizontal axis shows the compression factor. SC1 is the best-case coding and is stretched in the horizontal direction four times to give SC4. The example here shows a scenario where compressive sensing may work, i.e., SC4 has a 'win' region with respect to TC at highly compressed factors.
}
\label{fig:howtoi}
\vspace*{-3ex}
\end{figure}

{\bf Interpreting plots:} Refer to Figure \ref{fig:howtoi}.

\begin{figure}
\centering
\includegraphics[height=5.5cm]{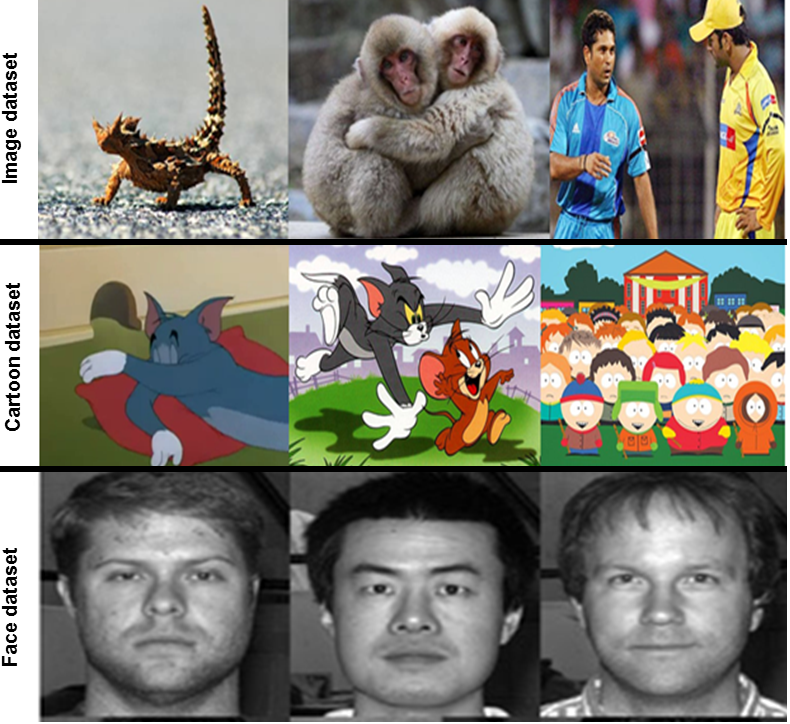}
\caption{ Image, Cartoon and face datasets: Image dataset made up of Flickr and internet images, Cartoon dataset: Tom and Jerry, Mickey mouse, Southpark, Face dataset: Yale B face dataset }
\label{fig: imagedataset}
\vspace*{-3ex}
\end{figure}
\begin{figure}
\centering
\includegraphics[width=7.5cm]{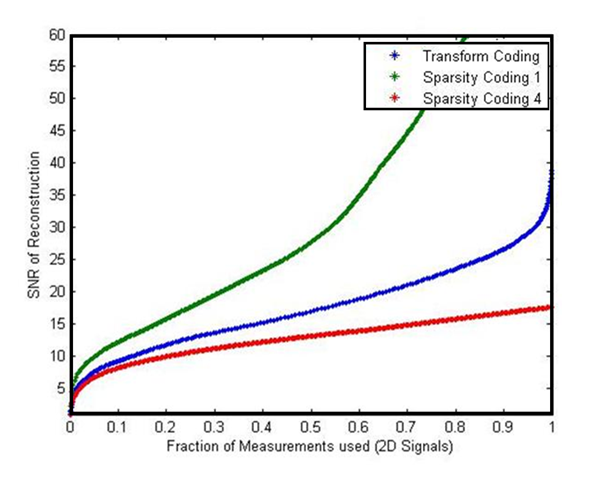}
\caption{ Compression of images gives low reconstruction SNR for lower values of C. Randomized sparsity coding (SC4) performs poorly in comparison with other methods}
\label{fig:Rand}
\vspace*{-3ex}
\end{figure}
\SubSection{Image datasets}
We performed the analysis of energy compaction and sparsity for images using 3 different kinds of image datasets: Random images, cartoons and Faces.

\textbf{Images:}
We compared the effectiveness using the following bases: DCT, FFT, Symlet, Daubechies and Haar wavelet.
Figure \ref{fig:Rand} shows the results.
Randomized sparse coding method becomes less effective as it does not know the non-zero coefficients locations.
Progressive transform coding performs better than randomized sparsity coding (SC4 ) for this dataset.

\begin{figure}
\centering
\includegraphics[width=8.5cm]{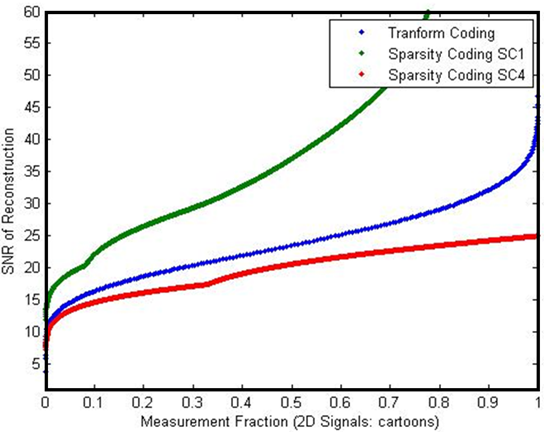}
\caption{ Although cartoons are sparse in gradient domain, TC continues to perform better than SC4.}
\label{fig:Cartoon}
\vspace*{-3ex}
\end{figure}

\textbf{Cartoon Images:} Figure \ref{fig:Cartoon} shows the results for experiments on random cartoon image dataset. Cartoons are sparse in gradient domain and have spectral distribution that is very different from real images.
This reflects in a better compression performance compared to the images dataset.

\begin{figure}
\centering
\includegraphics[width=7.5cm]{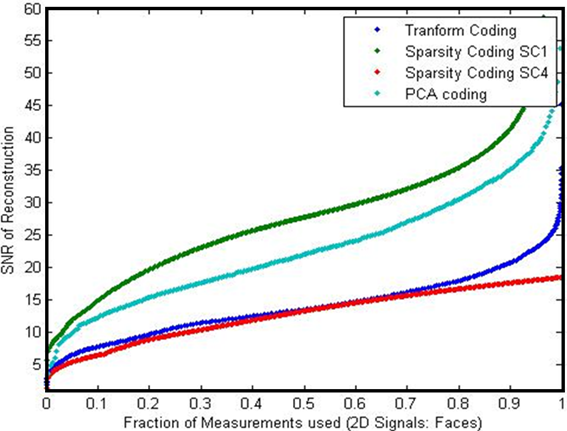}
\caption{PCA based compression performs well for faces. For other techniques they match with image compression performance}
\label{fig:Faces}
\vspace*{-3ex}
\end{figure}

\begin{figure}
\centering
\includegraphics[width=7.5cm]{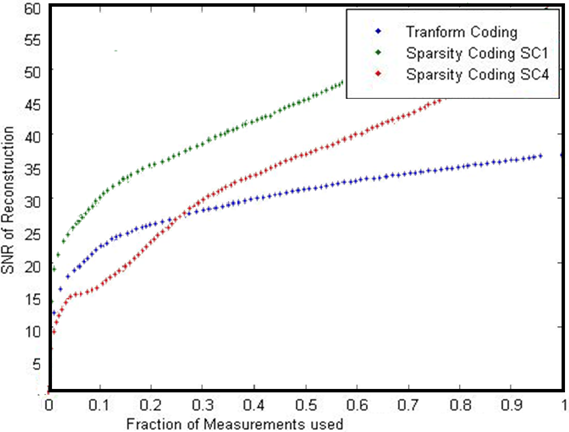}
\caption{Videos are sparser compared to images, due to inter-frame motion vector redundancies. Compression techniques perform better for videos than for images.}
\label{fig:Videos}
\vspace*{-3ex}
\end{figure}

\begin{figure}
\centering
\includegraphics[width=7.5cm]{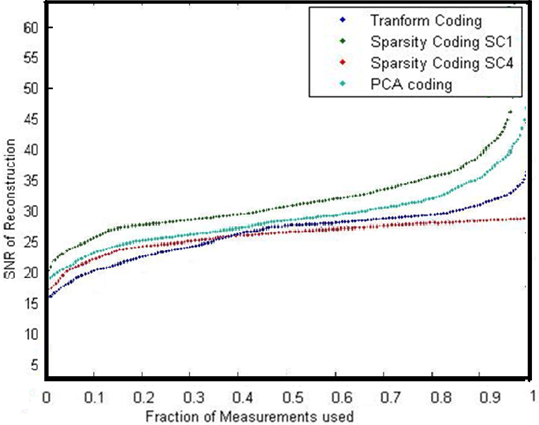}
\caption{Multispectral data shows intra-frame intensity variations. Compression performance is slightly better for multispectral data compared to images and videos. SC4 starts showing advantages at lower values of C}
\label{fig:MultiSpec}
\vspace*{-3ex}
\end{figure}

\begin{figure}
\centering
\includegraphics[width=7.5cm]{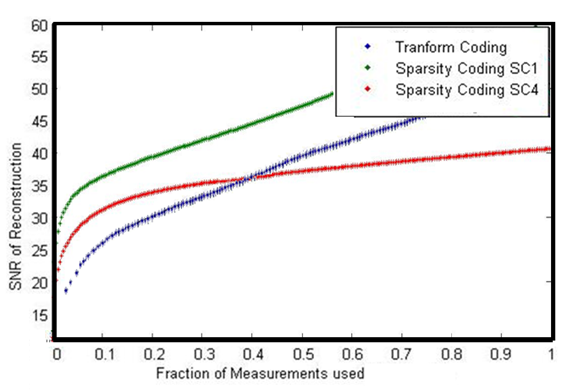}
\caption{Randomized Sparsity Coding (SC4) shows up to 5 dB better performance compared to Progressive Transform coding (TC). Light field sub-aperture views are sparser by nature and hence allow better reconstruction using SC4}
\label{fig:LFSpat}
\vspace*{-3ex}
\end{figure}

\textbf{Face Images:} For the face images, we performed the analysis using the data-independent bases as before (Figure \ref{fig:Faces}). But since faces have a lot of structural similarity we also performed analysis on face images using a data-dependent PCA basis. The basis was learnt from an independent face dataset and there was no overlap between the training and the testing datasets.
Data Dependant techniques perform well for face images.
Notice that while data-independent random sparse coding (SC4) and data-independent progressive transform coding (TC) seem to perform similarly, the data-dependent basis boosts performance significantly.
Nevertheless, there is not a great benefit in performing Randomized sparsity coding (SC4) over Progressive transform coding (TC) even for this dataset.
\SubSection{Videos}
Videos exhibit comparatively more redundancy because of inter-frame overlaps.
For videos SC4 does seem to perform marginally better than transform coding at very low compression factors (See Figure \ref{fig:Videos}).
Since most interesting compression techniques are concerned with larger compression factors, SC4 does not provide a significant advantage over TC even for such video data.
The performance of both these methods maybe improved significantly using motion compensation and other model based methods. The results presented in this paper do not extend directly to such model based methods.
\begin{figure}
\centering
\includegraphics[height=5.5cm]{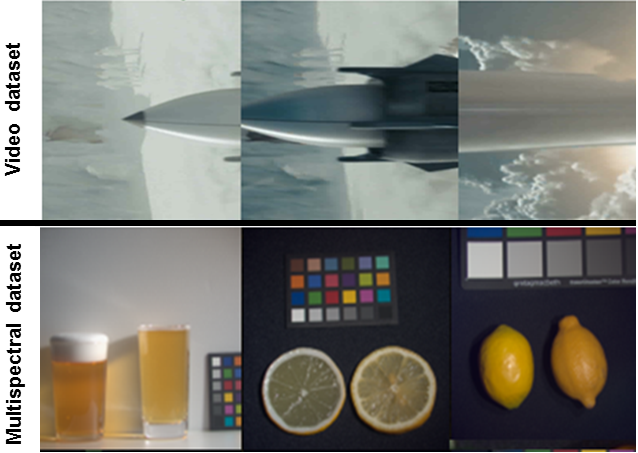}
\caption{ Video dataset- HD documentaries and movies, Multispectral dataset-CAVE labs, Columbia University}
\label{fig:videodataset}
\vspace*{-3ex}
\end{figure}
\SubSection{Multispectral}
Figure \ref{fig:MultiSpec} shows that multi-wavelength capture using Randomized Sparsity coding is beneficial. The intensity level variations among frames make the multispectral data sparser in third dimension. Therefore, employing sparsity based compression techniques lead to significant benefits at large compression factors.Since this is the point of operation that is practical, we see that compressive sampling confers a sampling advantage over traditional transform coding for this dataset.
\SubSection{Lightfields}
For the light-fields dataset, we notice in Figure \ref{fig:LFSpat} that Randomized Sparsity Coding (SC4) does perform significantly better than Progressive Transform Coding (TC).
This indicates that sparse representations and compressive sensing confer a significant benefit over traditional techniques for capture and representation of light-fields.

\begin{figure}
\centering
\includegraphics[width=7.5cm]{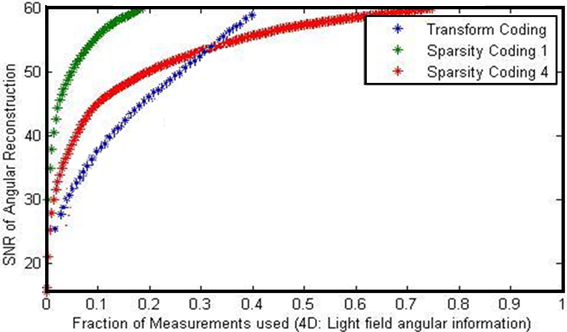}
\caption{ Light fields in angular domain are relatively sparse. The angular sampling of rays does not show significant pixel value variations. SC4 shows notable reconstruction SNR values at lower values of C. The overall reconstruction performance is best among the dataset types tested}
\label{fig:LFAng}
\vspace*{-3ex}
\end{figure}

The information contained in light-fields that is not available in traditional images is subtle disparity information and information about specular highlights.
The adjacent views of the light-field are usually very similar and the reconstruction SNR might be high even when this additional information (disparity etc) is not well captured.
In order to evaluate whether this information embedded in the light-field is well captured, we also evaluated another metric for light-fields.
We analyze the SNR of the center view subtracted Light-Field in order to evaluate the ability of these techniques to capture the angular information.
The results are shown in Figure \ref{fig:LFAng}.
This shows that SC4 performs better than TC even for this angular information metric.
Therefore, the disparity information is also better preserved by the Randomized sparsity coding techniques.
\begin{figure}
\centering
\includegraphics[width=7.5cm]{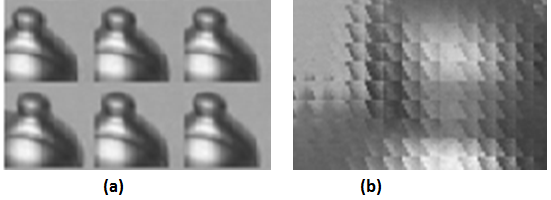}
\caption{Stanford Light field archive (a) Spatial (b) Angular disparity}
\label{fig:LFdataset}
\vspace*{-3ex}
\end{figure}

\Section{Conclusions}
We chose conditions favorable to projective signal capture, e.g., ignoring capture and reconstruction issues. Noise will be highly amplified after reconstruction. Randomized projections require basis pursuit algorithms assume near-idealized sensors and hence are highly susceptible to sensor noise or variation from system model. Quantization alone adds one-half of a digital level as noise (For an 8 bit image, the SNR of measured values is clamped at about 53dB.) Randomized projections are more likely to use high frequency patterns in optical path (e.g. coded apertures) making diffraction or calibration more critical. Bayesian inference and other prior based methods may benefit both types of projections.
Our discussion was limited to linear and non-adaptive projections. Results of the empirical experiments indicate that under idealized sensor and reconstruction conditions, randomized projections offer a benefit for acquiring high dimensional signals such as multi-spectral and light-field data in roughly this order.
\begin{center}{\small images $<$ faces $< $cartoons $<$ videos $<$ multispectral $<$ LF }\end{center}
Thus more research is needed in novel hardware for randomized projection based approaches for light fields and even reflectance fields.

The sparsity of visual signals is difficult to quantify. By experimenting over large sets, we hope to have captured the characteristics of natural visual signals. There is a remote possibility that the data is in fact significantly more sparse in some unusual yet-to-be-discovered transform basis. But that would be useful for general (non-linear, software based) image compression as well. The lack of such software algorithm indicates low likelihood of such a transform and we consider our empirical results as representative. Hence, researchers should further investigate non-linear or data-adaptive measurements - like use of learned dictionaries or over-complete dictionaries directly suitable for capture purposes.

\begin{figure}

\centering

\includegraphics[height=4.5cm]{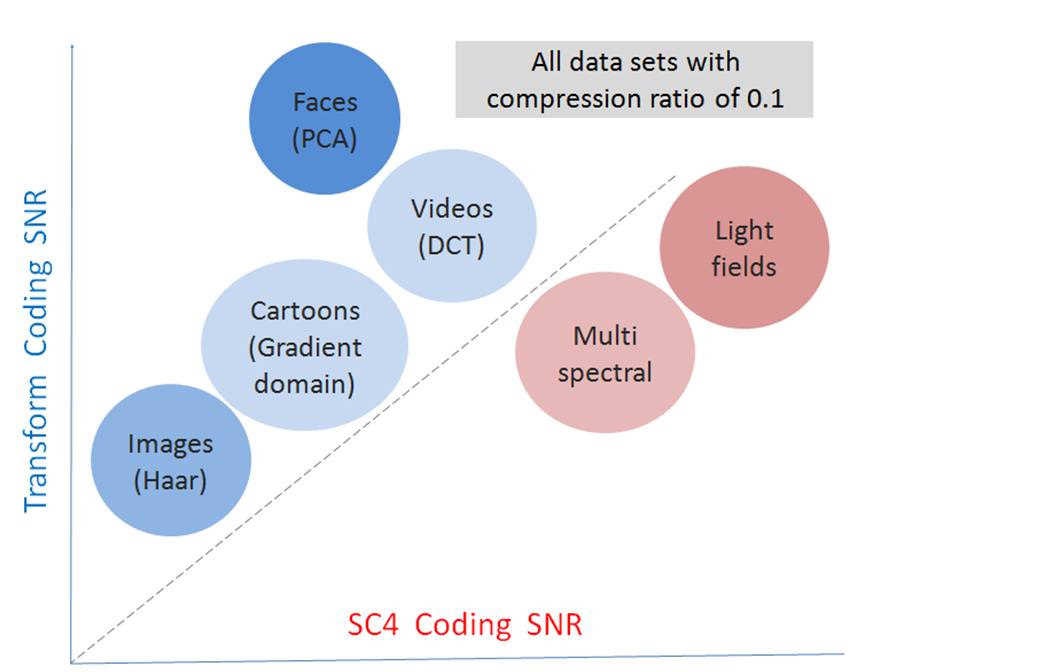}

\caption{ Summary of relative performance plotted as progressive transform coding SNR against  randomized sparsity coding SNR for each visual data type. Images have low SNR for both. Higher dimensional signals such as multispectral data and light fields have higher SNRs for SC4 coding.}

\label{fig:IndividualPerfPlots}

\vspace*{-3ex}

\end{figure}

{\small
\bibliographystyle{ieee}
\bibliography{ROHITICCPbib}

\begin{thebibliography}{10}\itemsep=-1pt

\bibitem{DBLP:journals/tsp/AbdelnourS05}
A.~F. Abdelnour and I.~W. Selesnick.
\newblock Symmetric nearly shift-invariant tight frame wavelets.
\newblock {\em IEEE Transactions on Signal Processing}, 53(1):231--239, 2005.

\bibitem{Ashok:10}
A.~Ashok and M.~A. Neifeld.
\newblock Compressive imaging: Hybrid projection design.
\newblock {\em Imaging Systems}, page IWD3, 2010.

\bibitem{Babacan:2009:667}
S.~D. Babacan, R.~Ansorge, M.~Luessi, R.~Molina, and A.~K. Katsaggelos.
\newblock Compressive sensing of light fields.
\newblock In {\em IEEE International Conference on Image Processing}, Cairo,
  Egypt, July 2009.

\bibitem{DBLP:journals/corr/abs-0808-3572}
R.~G. Baraniuk, V.~Cevher, M.~F. Duarte, and C.~Hegde.
\newblock Model-based compressive sensing.
\newblock {\em CoRR}, abs/0808.3572, 2008.

\bibitem{candes2006rup}
E.~Candes, J.~Romberg, and T.~Tao.
\newblock {Robust uncertainty principles: exact signal reconstruction from
  highly incomplete frequency information}.
\newblock {\em Information Theory, IEEE Transactions on}, 52(2):489--509, 2006.

\bibitem{2006ssr}
E.~Candes, J.~Romberg, and T.~Tao.
\newblock {Stable signal recovery from incomplete and inaccurate measurements}.
\newblock {\em COMMUNICATIONS ON PURE AND APPLIED MATHEMATICS}, 59(8):1207,
  2006.

\bibitem{537169}
R.~J. Clarke.
\newblock {\em Transform Coding of Images}.
\newblock Academic Press, Inc., Orlando, FL, USA, 1985.

\bibitem{WaveletTC}
B.~L.~B. DeVore, R.A.;~Jawerth.
\newblock Image compression through wavelet transform coding.
\newblock {\em IEEE Transactions on Information Theory}, 38(2):719--746, 1992.

\bibitem{DBLP:journals/tit/Donoho06}
D.~L. Donoho.
\newblock Compressed sensing.
\newblock {\em IEEE Transactions on Information Theory}, 52(4):1289--1306,
  2006.

\bibitem{Drori_compressedvideo}
I.~Drori.
\newblock Compressed video sensing.

\bibitem{Singlepixel}
M.~F.~D. et~al.
\newblock Single-pixel imaging via compressive sampling.
\newblock {\em IEEE Signal processing magazine}, 25(2):83--91, 2008.

\bibitem{Gehm:07}
M.~E. Gehm, R.~John, D.~J. Brady, R.~M. Willett, and T.~J. Schulz.
\newblock Single-shot compressive spectral imaging with a dual-disperser
  architecture.
\newblock {\em Opt. Express}, 15(21):14013--14027, 2007.

\bibitem{GeBeKr01}
A.~Georghiades, P.~Belhumeur, and D.~Kriegman.
\newblock From few to many: Illumination cone models for face recognition under
  variable lighting and pose.
\newblock {\em IEEE Trans. Pattern Anal. Mach. Intelligence}, 23(6):643--660,
  2001.

\bibitem{275888}
J.~Jeong.
\newblock The jpeg standard.
\newblock pages 91--99, 1997.

\bibitem{1553463}
J.~Mairal, F.~Bach, J.~Ponce, and G.~Sapiro.
\newblock Online dictionary learning for sparse coding.
\newblock pages 689--696, 2009.

\bibitem{1231494}
J.~F. Murray and K.~Kreutz-Delgado.
\newblock Learning sparse overcomplete codes for images.
\newblock {\em J. VLSI Signal Process. Syst.}, 46(1):1--13, 2007.

\bibitem{DBLP:journals/tog/PeersMLGMRD09}
P.~Peers, D.~Mahajan, B.~Lamond, A.~Ghosh, W.~Matusik, R.~Ramamoorthi, and
  P.~E. Debevec.
\newblock Compressive light transport sensing.
\newblock {\em ACM Trans. Graph.}, 28(1), 2009.

\bibitem{1291239}
F.~Pereira.
\newblock Mpeg multimedia standards: evolution and future developments.
\newblock In {\em MULTIMEDIA '07: Proceedings of the 15th international
  conference on Multimedia}, pages 8--9, New York, NY, USA, 2007. ACM.

\bibitem{1772035}
R.~Rubinstein, M.~Zibulevsky, and M.~Elad.
\newblock Double sparsity: learning sparse dictionaries for sparse signal
  approximation.
\newblock {\em Trans. Sig. Proc.}, 58(3):1553--1564, 2010.

\bibitem{Sen:CS_DualPhoto:2009}
P.~Sen and S.~Darabi.
\newblock {Compressive Dual Photography}.
\newblock {\em Computer Graphics Forum}, 28(2):609 -- 618, 2009.

\bibitem{Weiss_learningcompressed}
Y.~Weiss, H.~Chang, and W.~Freeman.
\newblock Learning compressed sensing.
\newblock 2007.

\bibitem{wright2009robust}
J.~Wright, A.~Yang, A.~Ganesh, S.~Sastry, and Y.~Ma.
\newblock {Robust face recognition via sparse representation}.
\newblock {\em IEEE Transactions on Pattern Analysis and Machine Intelligence},
  pages 210--227, 2009.

\end{thebibliography}
}

\end{document}